\documentclass{article}
\usepackage[a4paper,top=3cm,bottom=2cm,left=3cm,right=3cm,marginparwidth=1.75cm]{geometry}

\usepackage[utf8]{inputenc} % allow utf-8 input
\usepackage[T1]{fontenc}    % use 8-bit T1 fonts
\usepackage{hyperref}       % hyperlinks
\usepackage{url}            % simple URL typesetting
\usepackage{booktabs}       % professional-quality tables
\usepackage{amsfonts}       % blackboard math symbols
\usepackage{nicefrac}       % compact symbols for 1/2, etc.
\usepackage{microtype}      % microtypography
\usepackage{amssymb}
\usepackage{amsmath}
\usepackage{mathtools,mathrsfs}
\usepackage{blkarray}% http://ctan.org/pkg/blkarray

\usepackage{comment}
\title{Understanding overfitting peaks in generalization error: Analytical  risk curves for $l_2$ and $l_1$ penalized interpolation}

\author{
Partha P Mitra  \\
  Cold Spring Harbor Laboratory\\
  Cold Spring Harbor\\
  NY, NY 11724\\
  \texttt{mitra@cshl.edu} \\
}

\begin{document}

\maketitle

\begin{abstract}
Traditionally in regression one minimizes the number of fitting parameters or uses smoothing/regularization to trade training ({\it TE)} and generalization error ({\it GE}). Driving {\it TE} to zero by increasing fitting degrees of freedom ({\it dof}) is expected to increase {\it GE}. However modern big-data approaches, including deep nets, seem to over-parametrize and send {\it TE} to zero (data interpolation) without impacting {\it GE}. Over-parametrization has the benefit that global minima of the empirical loss function proliferate and become easier to find. These phenomena have drawn theoretical attention. Regression and classification algorithms have been shown that interpolate data but also generalize optimally. An interesting related phenomenon has been noted: the existence of non-monotonic risk curves, with a peak in {\it GE} with increasing {\it dof}. It was suggested that this peak separates a classical regime from a modern (interpolating) regime where over-parametrization improves performance. Similar over-fitting peaks were reported previously (cf. statistical physics approach to learning) and attributed to increased fitting model flexibility at the peak. We introduce a generative and fitting model pair ("Misparametrized Sparse Regression" or MiSpaR) and show that the overfitting peak can be dissociated from the point at which the fitting function gains enough {\it dof'}s to match the data generative model and thus provides good generalization. This complicates the interpretation of overfitting peaks as separating a "classical" from a "modern" regime. Data interpolation itself cannot guarantee good generalization: we need to study the interpolation with different penalty terms. We present analytical formulae for {\it GE} curves for MiSpaR with $l_2$ and $l_1$ penalties, in the interpolating limit (regularization parameter $\lambda\rightarrow 0$). These risk curves exhibit important differences and help elucidate the underlying phenomena.

\end{abstract}

\section{Introduction}
Modern machine learning has two salient characteristics: large numbers of measurements $m$, and non-linear parametric models with very many fitting parameters $p$, with both $m$ and $p$ in the range of $10^6-10^9$ for many applications. Fitting data with such large numbers of parameters stands in contrast to the inductive scientific process where models with small numbers of parameters are normative. Nevertheless, these large-parameter models are successful in dealing with real life complexity, raising interesting theoretical questions about the generalization ability of models with large numbers of parameters, particularly in the overparametrized regime $\mu=p/m>1$.  

Classical statistical procedures trade training (TE) and generalization error (GE) by controlling the model complexity. Sending TE to zero (for noisy data) is expected to increase GE\cite{james2013introduction}. However deep nets seem to over-parametrize and drive TE to zero (data interpolation) while maintaining good GE\cite{zhang2017understanding,belkin2018understand}. Over-parametrization has the benefit that global minima of the empirical loss function proliferate and become easier to find\cite{ma2017interpolation,mitra2018fast}. These observations have led to recent theoretical activity\cite{belkin2018overfitting,belkin2018understand,liang2018just}. Regression and classification algorithms have been shown that interpolate data but also generalize optimally\cite{belkin2018overfitting}. An interesting related phenomenon has been noted: the existence of a peak in GE with increasing fitting model complexity\cite{belkin2018reconciling,advani2017high,geiger2018jamming,hastie2019surprises}. In \cite{belkin2018reconciling} it was suggested that this peak separates a classical regime from a modern (interpolating) regime where over-parametrization improves performance. While the presence of a peak in the GE curve is in stark contrast with the classical statistical folk wisdom where the GE curve is thought to be U-shaped, understanding the significance of such peaks is an open question, and motivates the current paper. Parenthetically, similar over-fitting peaks were reported almost twenty years ago (cf. statistical physics approach to learning) and attributed to increased fitting model entropy near the peak (see in particular Figs 4.3 and 5.2 in \cite{engel2001statistical}). 

\begin{figure}
  \centering
    \includegraphics[width=\linewidth]{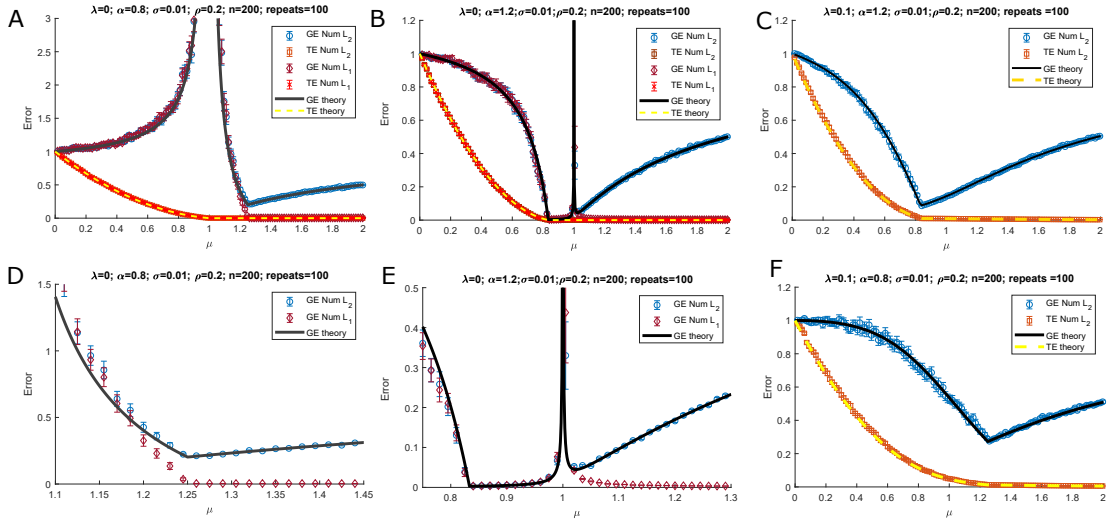}
   
    \caption{Numerical simulations of the MiSpaR model inferred using $l_2$ and $l_1$ penalties are compared with theoretical TE and GE curves for $l_2$ regularized regression. (A,B) and zooms (D,E) correspond to the interpolation limit $\lambda\rightarrow 0$. Plots in (C,F) show a theory-simulation comparison just for the $l_2$ case with $\lambda=0.1$. Here $n=200$, and the numerical values are averaged over 100 draws of the design matrix $X$, parameters $\beta$ and measurement noise $\sigma$. The rows of the design matrix are sub-sampled in the $\mu<1$ regime. Standard errors across the 100 trials are shown. Note that for $\mu\alpha>1$, the GE values for the $l_1$ case are close to zero, whereas the values for the $l_2$ penalized case can be much larger. Note also that the overfitting peak is much larger for $\alpha<1$ than for $\alpha>1$, and that the region of good generalization starts at $\mu=1/\alpha$, which can be to the left or right of the overfitting peak depending on the value of the undersampling parameter $\alpha$. For a {\it single} draw of the design matrix $X$, one still obtains agreement between the theoretical curves and simulations due to self-averaging, although there is greater scatter (Fig.2).}
\end{figure}

\subsection{Summary of Results}
\begin{enumerate}
    \item We introduce a model, Misparametrized (or Misspecified) Sparse Regression (MiSpaR), which separates the number of measurements $m$, the number of model parameters $n$ (which can be controlled for sparsity by a parameter $\rho$), and the number of fitting degrees of freedom $p$. \footnote{A similar misspecified model has been studied in \cite{hastie2019surprises} with $l_2$ regularization, but this paper did not study the effects of sparsity and $l_1$ penalized regression.}
    \item We obtain analytical expressions for the GE and TE curves for $l_2$ penalized regression in the "high-dimensional" asymptotic regime $m,p,n\rightarrow \infty$ keeping the ratios $\mu=m/p$ and $\alpha=m/n$ fixed. We also present analytical expressions that permit computation of the GE for $l_1$ penalized regression, and obtain explicit expressions for $\mu<1$ and $\mu>>1$ for the interpolating limit $\lambda\rightarrow 0$.
    \item We show that for $\lambda\rightarrow 0$ and for $\sigma>0$, the overfitting peak appears at the data interpolation point $\mu=1$ ($p=m$) for both $l_2$ and $l_1$ penalized interpolation ($GE\sim|1-\mu|^{-1}$ near $\mu=1$), but does not demarcate the point at which "good generalization" first occurs, which for small $\sigma$ corresponds to the point $p=n$ ($\mu\alpha=1$) (Figures 1-3). The region of good generalization can start before or after the overfitting peak. The overfitting peak is suppressed at finite values of lambda.
    \item For infinitely large overparametrization, generalization does not occur: $GE(\mu\rightarrow\infty)=1$ for both $l_2$ and $l_1$ penalized interpolation. However, for small values of the sparsity parameter $\rho$ and measurement noise variance $\sigma^2$, there is a large range of values of $\mu$ where $l_1$ regularized interpolation generalizes well, but $l_2$ penalized interpolation generalizes poorly (Figure 1). This range is given by $1<<log(\mu)<<\frac{1}{\sigma^2},\frac{1}{\rho}$, with $\sigma^2,\rho/\alpha<<1$. 
    
    The reason for this difference is that in this regime the sparsity penalty is effective, and suppresses noise-driven mis-estimation of parameters for the $l_1$ penalty. This concretely demonstrates how generalization properties of penalized interpolation depend strongly on the inductive bias, and are not properties of data interpolation {\it per se}.
    
    \item For $\sigma=0$ and for $\mu>1$, $GE(l_2)>0$. In contrast, if $\alpha$ is greater than a critical value $\alpha_c(\rho)$ that depends on $\rho$, then $GE_1=GE(l_1)=0$ for a range of overparameterization $\frac{1}{\alpha}\leq\mu\leq\mu_c$. The maximum overparametrization $\mu_c$ for which $GE_1=0$ depends on $\frac{\rho}{\alpha}$. For small values of $\frac{\rho}{\alpha}$,  $\mu_c\sim\sqrt{\frac{\pi\alpha}{2\rho}}e^{\frac{\alpha}{2\rho}}$. For $\mu>\mu_c$, $GE(l_1)$ rises quadratically from zero ($GE_2(\mu>\mu_c)\propto (\mu-\mu_c)^2$ for small $\mu-\mu_c$) and $GE_1(\mu\rightarrow\infty)=1$. 
    
    \item For $\sigma=0$ and $\alpha>\alpha_c(\rho)$, $GE_1$ goes to zero linearly at $\mu\alpha=1$ ($GE_1\propto(\frac{1}{\alpha}-\mu)$ for $\frac{1}{\alpha}-\mu$ small). When $\alpha=\alpha_c(\rho)$, $GE_1=0$ only at the single point $\mu_c=\frac{1}{\alpha_c(\rho)}$. In this case $GE_1$ goes to zero with a nontrivial $\frac{2}{3}$ power $GE_1(\mu\lesssim \frac{1}{\alpha_c(\rho)})\propto (\frac{1}{\alpha_c(\rho)}-\mu)^\frac{2}{3}$ on the left, but rises quadratically on the right $GE_1(\mu\gtrsim \frac{1}{\alpha_c(\rho)})\propto(\frac{1}{\mu-\alpha_c(\rho)})^2$. For $\alpha<\alpha_c(\rho)$, $GE_1>0$ for all values of $\mu$.  
\end{enumerate}

\section{Model: Misparametrized Sparse Regression}
Usually in linear regression the same (known) design matrix $x_{ij}$ is used both for data generation and for parameter inference. In MiSpaR the generative model has a fixed number $n$ of parameters $\beta_j$, which generate $m$ measurements $y_i$, but the number of parameters $p$ in the inference model is allowed to vary freely, with $p<n$ corresponding the under-parametrized and $p>n$ the over-parametrized case. For the under-parametrized case, a truncated version of the design matrix is used for inference, whereas for the over-parametrized case, the design matrix is augmented with extra rows. 

In addition, we assume that the parameters in the generative model are sparse, and consider the effect of sparsity-inducing regularization in the interpolation limit. Combining misparametrization with sparsity is important to our study for two reasons
\begin{itemize}
    \item Dissociating data interpolation (which happens when $\mu=1$, $\lambda\rightarrow 0$) from the regime where good generalization can occur (this is controlled by the undersampling $\alpha$ as well as by the model sparsity $\rho$).
    \item We are able to study the effect of different regularization procedures on data interpolation in an analytically tractable manner and obtain analytical expressions for the generalization error. 
\end{itemize}
    
To motivate this separation and the sparsity constraint, consider the following hypothetical scenario: suppose we are trying to make a diagnostic prediction characterized by some scalar parameter $y_i$ for the $i^{th}$ individual. We obtain a training sample of $n$ individuals, and for each individual measure a set of biological variables $x_{ij}$ with $j=1..p$ ({\it e.g.} age, weight, etc) which we think might have diagnostic predictive value. We then proceed to fit a predictive model using these phenotypic variables (which will themselves stochastically vary from person to person, but can be measured for a new "test" person). 

Now consider that the generative model is itself linear, but (i) only $n$ of these parameters have any effect on the diagnoses, and (ii) that there are latent or unmeasured variables that further subdivides the population into groups where only $k$ of the $n$ parameters have predictive value, with a different choice of the $k$ parameters in each group. For example, in one such latent group height may have predictive value, and not in another. Neither $n$, nor the specific parameters which could have non-zero values, are known in advance (and one might err on the side of caution in measuring extra phenotypic variables that may or may not impact the disease in question), although one might be able to prioritize some of the parameters due to prior scientific knowledge. Since we can now obtain data from large populations, it is tempting to measure a large number of biological variables per person - this is almost inevitable given low-cost genomics and advanced imaging techniques. Thus, the model studied here is well-motivated and relates to real life scenarios.

\begin{comment}
\begin{figure}
  \centering
\includegraphics[width=\linewidth]{composite.eps}
%\includegraphics{composite.eps}
    \caption{Numerical simulations of the MiSpaR model inferred using $l_2$ and $l_1$ penalties are compared with theoretical TE and GE curves for $l_2$ regularized regression. (A,B) and zooms (D,E) correspond to the interpolation limit $\lambda\rightarrow 0$, whereas (C) shows a theory-simulation comparison just for the $l_2$ case with $\lambda=0.5$. Here $n=300$, and the simulations correspond to a single draw of the design matrix $X$, parameters $\beta$ and measurement noise $\sigma$ which is sub-sampled in the $\mu<1$ regime (thus theory-numerical correspondence relies on self-averaging properties). Note that for $\mu>1$, the GE values for the $l_1$ case are close to zero, whereas the values for the $l_2$ penalized case are much larger. Note also that the overfitting peak is much larger for $\alpha<1$ than for $\alpha>1$, and that the region of good generalization starts at $\mu=1/\alpha$, which can be to the left or right of the overfitting peak depending on the value of the undersampling parameter $\alpha$. Note that for a single draw of the design matrix $X$, th
}
\end{figure}
\end{comment}

\begin{figure}
  \centering
    \includegraphics[width=\linewidth]{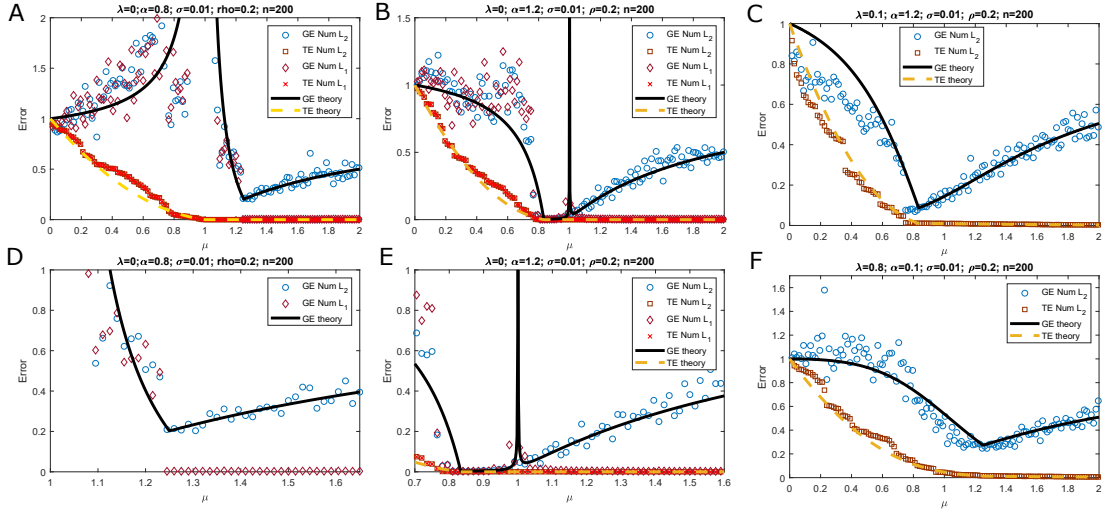}
    \caption{A simulation with a single draw of $X$, with parameters otherwise corresponding to Fig.1. Although considerable scatter is seen, there is qualitative correspondence between theory and simulation. Averaging over 100 such draws produces a better correspondence (Fig.1)}
\end{figure}

\subsection{Generative Model}
We assume that the (known/measured) design variables are i.i.d. Gaussian distributed from one realization of the generative model to another \footnote{Note that these choices are convenient, but could be relaxed. The calculations rely on the large-n asymptotics of random matrix theory, and therefore exhibit corresponding universality properties.}  with variance $1/n$. This choice of variance is important to fix normalization. Other choices have been also employed in the literature (notably $x_{ij}\sim N(0,1/m)$) - this is important to keep in mind when comparing with literature formulae where factors of $\alpha$ may need to be inserted appropriately to obtain a match. 
        \begin{align}
        & y_{i}=\sum_{j=1}^{n} x_{i j} \beta_{j}+n_{i} \nonumber \\
        & n_{i} \sim N(0, \sigma) \quad \quad \beta_{j} \sim(1-p) \delta_{\beta, 0}+\rho\pi(\beta)\nonumber \\
        & \pi(\beta)\sim N(0,1) ~~\text{unless~otherwise~specified}\nonumber \\
        & x_{i j} \sim N\left(0, \frac{1}{n}\right)\quad \quad 
        i=1 \dots m \text{~measurements}  \quad
        j=1 \dots n \text{~generative~parameters} \nonumber \\
        & \text{Undersampling:~} \alpha=m/n \quad \quad \text{Sparsity:~} \rho \quad \quad \text{Overparametrization:~} \mu=p/m \nonumber
     \end{align}
Here $\pi(\beta)$ is the distribution of the non-zero model parameters. We assume this distribution to be Gaussian as this permits closed form evaluation of integrals appearing in the $l_1$ case. Note that we term $\mu=p/m$ as overpametrization (referring to the case where $\mu>1$) and we term $\alpha=m/n$ as undersampling (referring to the case where $\alpha<1$).

\subsection{Inference Model}
The design matrix used for inference is mis-parametrized or mis-specified: under-specified (or partially observed) when $\mu\alpha<1 \equiv p<n$; over-specified, with extra, effect-free rows in the design matrix when $\mu\alpha>1 \equiv p>n$\\
\begin{align}
   & x^{inf}_{ij}=x_{ij}, \quad \quad & j=1 \dots p  \quad \quad \text{~if~}\ p\leq n \nonumber \\
   & x^{inf}_{ij}=x_{ij}^{extra}, \quad \quad  & j=n+1 \dots p \quad \text{~if~}\ p>n \nonumber
\end{align} 
In the remaining sections, we will generally not explicitly annotate $X^{inf}$ for the design matrix used in inference, since the usage is clear from context. Parameter inference is carried out by minimizing a penalized mean squared error\\  
    \begin{equation}\hat{\beta} = argmin_{\beta} \frac{1}{2}\left|Y-X^{inf}\beta\right|^{2}+\lambda V\left(\beta\right)
    \nonumber
    \end{equation}
    Note that for $p>n$, the model parameters $\beta$ are augmented by $n-p$ zero entries. We consider $l_2$ and $l_1$ penalties (correspondingly 
    $V(\beta)=\frac{1}{2}|\beta|_{2}^{2}$ or $|\beta|_{1}$) and the interpolation limit is obtained by taking $\lambda\rightarrow 0$. For the $l_2$ penalty (ridge regression), $\hat{\beta_2} = (X^+X+\lambda)^{-1}X^+Y $. The training and generalization errors are defined as the normalized MSEs on training and test sets: 
    
    \begin{align}
    & \text{Training~Error~} (TE)=\frac{E\left|Y-X^{inf}\hat{\beta}\right|^{2}}{E|Y|^{2}} \nonumber \\ & \text{Generalization~Error~} (GE)=\frac{E\left|Y^{New}-X^{inf}_{new}\hat{\beta}\right|^{2}}{E|Y^{new}|^{2}} \nonumber \end{align}. 
    Note that the expectation $E$ is taken simultaneously over the parameter values, the design matrix and measurement noise. Where necessary below for clarity, we explicitly separate out the averaging over the design matrix. 

\section{Risk Curves}

The focus in the paper is to obtain exact analytical expressions for the risk, rather than bounds. We do this in the limit where $n,p,m$ all tend to infinity, but the ratios $\alpha=m/n,~\mu=p/m$ are held finite. Similar "thermodynamic" or "high-dimensional" limiting procedures are used in statistical physics, {\it eg} in the study of random matrices and spin-glass models in large spatial dimensions\cite{mezard1987spin,mehta2004random}. Such limits are also well-studied in modern statistics\cite{wainwright2019high} (for example to understand phase-transition phenomena in the LASSO algorithm\cite{DonohoPT}). 

In this limit, we give analytical formulae for TE and GE with $l_2$ or ridge regularization. For $l_1$ regularization, explicit formulae are given in some parameter regimes. More generally for the $l_1$ case we obtain a pair of simultaneous nonlinear equations in two variables, which can be solved numerically to obtain the GE. The nonlinear equations are given in closed form without hidden parameters and do not require integration.  

\subsection{$l_2$ Risk Curves: Formulae}

The formulae are split into two cases: the underspecified case $p\leq n$ $(\mu \alpha\leq 1)$, and the overspecified case where $p\geq n$ $(\mu \alpha\geq 1)$. In the former case, there are un-observed columns of the design matrix which correspond to generative model parameters that can be non-zero, but are missing from the fitting model. Due to the model setup, different parameters are non-zero for different measurements (recall the motivating example). Thus, these un-observed parameters contribute an effective additive noise, resulting in an increase in the effective noise variance to 
\begin{equation}
\sigma_{eff}^2=\sigma^2+(1-\mu\alpha)\rho 
\nonumber \end{equation}
    In the overspecified case, there are extra rows in the design matrix, which correspond to parameters in the generative model which are always zero. Since these parameters appear in the fitting model, they can in general have non-zero inferred values (due to measurement noise), and contribute to the estimation variance and generalization error. 

\subsubsection{{\it Underspecified case}: $p\leq n$ $(\mu \alpha\leq 1)$} $\sigma_{eff}^2=\sigma^2+(1-\mu\alpha)\rho$. The training error is given by the following formulae:

\begin{align}
& TE = \frac{\sigma_{eff}^2}{\sigma^2+\rho} \left[ (1-\mu) \theta(1-\mu)  +\frac{1}{2}\left\{({\cal A}-|1-\mu|)-\frac{\lambda^2\rho}{\alpha\sigma_{eff}^2}+\frac{\lambda}{\alpha {\cal A}}\left(\frac{\rho\lambda}{\sigma_{eff}^2}-1\right)\left(\frac{\lambda}{\alpha}+(1+\mu)\right)\right\}\right]
\nonumber \\
&   {\cal A}=\left[\left(\frac{\lambda}{\alpha}+\mu_{+}\right)\left(\frac{\lambda}{\alpha}+\mu_-\right)\right]^{1 / 2};  \quad \mu_{\pm}=(1 \pm \sqrt{\mu})^{2}\nonumber
\end{align}

The generalization error is given by the following formula: 

\begin{equation}
GE =\frac{1}{\sigma^{2}+\rho}\Bigl[\mu\alpha\rho \theta(\mu-1)\left(1-\frac{1}{\mu}\right)+\frac{1}{2}\left\{\rho \alpha({\cal A}-|1-\mu|)+ 
    \frac{1}{{\cal A}}\left(\sigma_{eff}^2-\lambda\rho\right)\left(\frac{\lambda}{\alpha}+1+\mu\right)+\sigma_{eff}^2\right\}\Bigr] 
    \nonumber
\end{equation}

\begin{figure}
  \centering
    \includegraphics[width=0.7\linewidth]{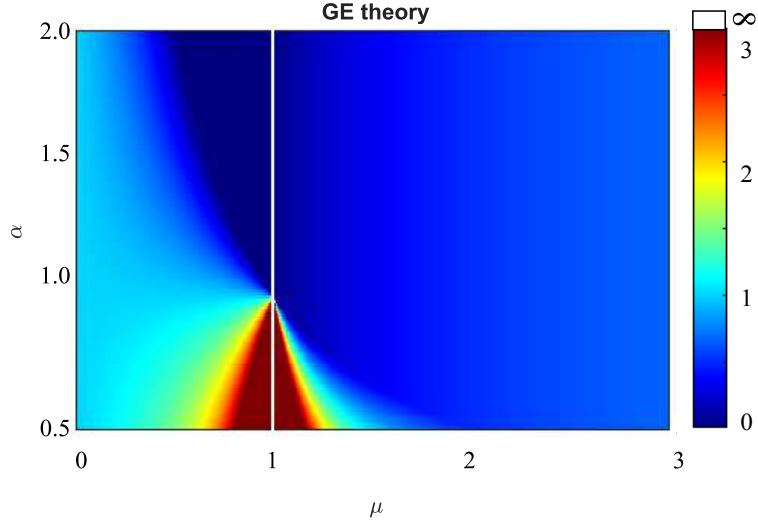}
    \caption{The theoretical generalization error for $l_2$ penalized interpolating regression is shown as a function of $\mu$ and $\alpha$ with substantial sparsity and small additive noise ($\rho=0.2,\sigma=0.01$). The noise peak at $\mu = 1$ where $GE_2=\infty$ appears as a vertical white line. It can be clearly seen that the starting point of the "good generalization" regime is dissociated from the data interpolation line at $\mu=1$ and starts at $\mu\alpha=1$ (corresponding to a parabolic curve which is visible in the figure, to the left of which one obtains small values of $GE_2$).}
\end{figure}

\subsubsection{{\it Overspecified case:} $p\geq n$ $(\mu \alpha\geq 1)$}

The formulae for TE and GE for the overspecified case, are obtained from the corresponding formulae above for the underspecified case, by making the substitutions $\sigma_{eff}^2 \rightarrow \sigma^2$ and $\rho \rightarrow \rho/(\mu \alpha)$. That this should be the case is intuitively obvious, since in the overspecified case there are no unobserved parameters contributing to an effective noise, on the other hand the model parameters are effectively more sparse when compared to the total number of fitting paramters, and it makes sense that $k/n$ should be replaced by $k/p$. This follows from the relevant derivation. We will not explicitly write out these formulae since they are straightforward to obtain by making the substitution mentioned. 

Note that in both cases (overspecified and underspecified), both TE and GE $\rightarrow 1$ when $\lambda \rightarrow \infty$, as can be verified by taking the limit in the corresponding formulae above. 

\subsubsection{\it Interpolating limit $\lambda \rightarrow 0$}

It is useful to collect together the formulae for TE and GE in the interpolating limit $\lambda\rightarrow 0$

\begin{eqnarray}
&TE(\lambda\rightarrow 0,l_2)=&\frac{\sigma^2+\rho (1-\mu\alpha)\theta(1-\mu\alpha)}{\sigma^{2}+\rho}\Bigl[(1-\mu) \theta(1-\mu)\Bigr]
\nonumber\\
&GE(\lambda\rightarrow 0,l_2) =&\frac{1}{\sigma^{2}+\rho}\Bigl[\rho\bigl\{\mu\alpha\theta(1-\mu\alpha)+\theta(\mu\alpha-1)\bigr\} \theta(\mu-1)\left(1-\frac{1}{\mu}\right)+\nonumber\\
& &(\sigma^2+\rho (1-\mu\alpha)\theta(1-\mu\alpha))\left\{\frac{\theta(1-\mu)}{1-\mu}+\frac{\mu\theta(\mu-1)}{\mu-1}\right\}\Bigr]
\nonumber
\end{eqnarray}

\subsection{$l_2$ Risk Curves: Derivation}

First consider the case $ p \leq n \quad \mu \alpha \leq 1 $. The design matrix $X$ can be split into two parts, \\
      $X = [X_0^{m \times p} X_u^{m\times(n-p)}]$ where the second part corresponds to $n-p$ parameters that are not in the fitting model. In this case, $ Y = X_0\beta_0 + N_{eff} $ with the effective noise being given by 
      \begin{equation} 
      n_{eff}^i = \sum_{j=p+1}^n x_{ij}\beta_j + n_i \nonumber
      \end{equation}
      Across different realizations of the generative model, the parameters $\beta_j$ vary with variance $ V(\beta_j) = \rho$. Thus the only change from ordinary ridge regression with $p$ parameters in the fitting model, is the replacement of the noise variance with an effective variance 
      \begin{equation}
V(n^{eff}_i) =\sigma_{eff}^2 = \sigma^2 + \rho(\frac{n-p}{n}) = \sigma^2 + \rho(1- \mu \alpha) \nonumber
\end{equation}
The inferred parameter values $\hat{\beta_0}$ are given by 
\begin{equation}
\hat{\beta_0} = (X_0^+X_0 + \lambda)^{-1}X_0^+\{X_0\beta_0 + N_{eff}\}
\nonumber
\end{equation}

A little algebra then shows that the training error is given by ($\lambda_i$ are the eigenvalues of the $p\times p$ Wishart matrix $X_0^{\dagger}X_0$)
  
  \begin{equation}
      \frac{E\left|Y-X_{0} \hat{\beta}_{0}\right|^2}{E|Y|^2}=\frac{1}{m\left(\sigma^{2}+\rho\right)}\bigg\{\sigma_{eff}^{2}(m-p)+\lambda^{2} E \sum_{i=1}^{p} \frac{\rho \lambda_{j}+\sigma_{eff}^{2}}{\left(\lambda_{j}+\lambda\right)^{2}}\bigg\}   
  \end{equation}

Under the assumptions (including the asymptotic limits) the eigenvalues $\lambda_i$ follow the Marchenko-Pastur distribution\cite{marchenko1967distribution} across different realizations of the generative model. Importantly, sums such as in Eq.1 are self-averaging, and even for a given realization of the generative model, can be replaced by the ensemble average for large enough n. The following formula is used to compute the necessary sums for the TE and GE (using the appropriate form of $f(\lambda)$ from the corresponding cases) 

\begin{equation}
    \frac{1}{p}\sum_{i=1}^{p} f(\lambda_i) = f(0)\theta(\mu-1)(1-\frac{1}{\mu}) + \frac{1}{2\pi}\int_{\mu_-}^{\mu_+}\frac{\sqrt{(\mu_+ - z)(z - \mu_-)}}{\mu z}f(\alpha z)dz
\end{equation}

Here $\mu_{pm}=(1\pm\sqrt{\mu})^2$. Applying this formula to the expression for the training error and performing the relevant integral using the method of residues and contour integration, one obtains the formula presented earlier for the training error for $\mu\alpha<1$. 

%A couple of points about calculating the contour integral: residues at the points $z=0$ and $z=-\lambda/\alpha$ as well as the residue at infinity have to be taken into account, and a convenient choice of the branch cut for the square root connects the points $z=\mu_-$ and $z=\mu_+$. 

To compute generalization error, one has to pick a new row $x_{new}$ of the design matrix. Only a subset $p<n$ of the parameters are used in the forward prediction, corresponding to a restricted portion $x^0_{new}$ of the vector $x_{new}$\\
\begin{align}
y_{new}&={x}_{new} \cdot {\beta}+n_{new}\nonumber\\
y_{fit}&= {x}_{new}^{0} \cdot \hat{\beta}_{0}\nonumber
\end{align}
 Some algebra then leads to the following equation for GE

\begin{equation}
    GE = \frac{1}{\sigma^{2}+\rho}\left[\sigma_{eff}^{2}+\frac{1}{n}E\sum_{j=1}^{p} \frac{\rho\lambda^{2}+\sigma_{eff}^{2}\lambda_{j}}{\left(\lambda+\lambda_{j}\right)^{2}}\right]
\end{equation}

Noting that $p/n=\mu\alpha$ and applying the Marchenko-Pastur distribution as above to compute the sum in the large $m,n,p$ limit, leads to the GE formula given in the previous section for $\mu\alpha<1$. The formulae for $\mu\alpha>1$ follow a very similar derivation.

\subsection{$l_1$ Risk Curves: Formulae}

The formulae in this section are for the case $\pi(\beta)\equiv N(0,1)$. Some other forms of $\pi(\beta)$ also lead to closed form expressions. The considerations of this section will continue to hold qualititatively as long as $\pi(0)>0$. Even better generalization is expected if $\pi(0)=0$, and especially if the distribution $\pi(\beta)$ has a gap region near zero where it is zero throughout. However, these other choices of $\pi(\beta)$ do not change the range of values of $\mu$ for which $GE_1(\sigma=0,\lambda\rightarrow 0)=0$. 

\subsubsection{{\it Underspecified case}: $p\leq n$ $(\mu \alpha\leq 1)$; $\sigma_{eff}^2=\sigma^2+(1-\mu\alpha)\rho$}

The generalization error with $l_1$ regularization is given by \\
\begin{equation}
GE_1=\frac{\alpha\sigma_{\xi}^2}{\sigma^2+\rho} \nonumber   
\end{equation}
Where the variable $\sigma_{\xi}$ has to be found by solving the following three equations Eq.4-6 with three unknowns $\tau$, $\hat{\rho}$ and $\sigma_{\xi}$. Note that $\hat{\rho}$ is the fraction of estimated parameters which are non-zero, this is a number which is constrained to be $0\leq \hat{\rho} \leq 1$. Also $\tau\geq 0$.

\begin{equation}
     1-\hat{\rho}=(1-\rho) \operatorname{Erf}\left(\frac{\tau}{\sqrt{2}}\right)+\rho \operatorname{Erf}\left(\frac{a}{\sqrt{2}}\right)    
\end{equation}

\begin{equation}
\sigma_{\xi}^{2}=\frac{1}{\alpha}\left(\sigma_{eff}^{2}+\mu\alpha\sigma_{\xi}^2[(1-\rho) A+\rho B]\right)
\end{equation}

\begin{equation}
    \frac{\lambda}{\alpha}=\tau \sigma_{\xi}(1-\mu \hat{\rho})
\end{equation}

where 
\begin{align}
&A=\left(1+\tau^{2}\right) \operatorname{Erfc}\left(\frac{\tau}{\sqrt{2}}\right)-\sqrt{\frac{2}{\pi}}\tau e^{-(\tau^2 / 2) }  \nonumber\\
&B=1+\tau^{2}-C_{0}\left\{\sqrt{\frac{2}{\pi}} a e^{- {a^2 / 2}}+\left(a^{2}-1\right)
\operatorname{Erf}\left(\frac{a}{\sqrt{2}}\right)\right\}-2 \operatorname{Erf}\left(\frac{a}{\sqrt{2}}\right)\nonumber
\end{align}

with $a=\tau\sigma_{\xi}/\sqrt{1+\sigma_{\xi}^2}$ and $C_0=1+1/\sigma_{\xi}^2$. $Erf(\tau)$ is the standard error function $Erf(\tau) = \frac{2}{\sqrt{\pi}}\int_0^\tau e^{-x^2} dx$

\subsubsection{{\it Overspecified case}: $p\geq n$ $(\mu \alpha\geq 1)$; $\sigma_{eff}^2=\sigma^2+(1-\mu\alpha)\rho$}

As for $l_2$ regularization, the formulae for the overspecified case can also be obtained by making the substitutions $\sigma_{eff}\rightarrow \sigma$ and $\rho \rightarrow \rho/(\mu \alpha)$. This involves only the first two equations above (the other equations remain the same): 
\begin{equation}
     1-\hat{\rho}=(1-\frac{\rho}{\mu\alpha}) \operatorname{Erf}\left(\frac{\tau}{\sqrt{2}}\right)+\frac{\rho}{\mu\alpha} \operatorname{Erf}\left(\frac{a}{\sqrt{2}}\right)    
\end{equation}
\begin{equation}
\sigma_{\xi}^{2}=\frac{1}{\alpha}\left(\sigma^2+\mu\alpha\sigma_{\xi}^2[(1-\frac{\rho}{\mu\alpha}) A+\frac{\rho}{\mu\alpha} B]\right)
\end{equation}

\subsubsection{Analytical expressions for GE when $\sigma>0$} 

One can obtain analytical insights by considering special cases and limits. In the interpolating limit $\lambda\rightarrow 0$, Eq.4 implies that either $\tau$ or $\sigma_{\xi}$ or $(1-\mu\hat{\rho})$ must go to zero. In order for $\sigma_{\xi}$ to go to zero, from Eq.6 it follows that $\sigma=0$ (noise-free case). This is an interesting limit as it corresponds to the well known algorithmic phase transition for $l_1$ penalized regression\cite{DonohoPT}. We will consider it in the next section, but first we but examine the finite noise case. 

For the considerations below, except where explicitly noted, we assume $\sigma>0$, which also implies from Eqs.5,8 that $\sigma_{\xi}>0$. Thus in the interpolating limit $\lambda\rightarrow 0$ one must either have $\tau\rightarrow 0 $ or $\hat{\rho}\rightarrow \frac{1}{\mu}$. We will consider these two cases in turn: as can be seen below, the first case corresponds to $\mu<1$ and the second case to $\mu>1$. 

{\bf Case 1: $\tau\rightarrow 0$ $\implies \mu<1$} \\
In this case, from Eq.4 or Eq.7 it follows that $\hat{\rho}=1$, ie all the fitted parameters are non-zero. It can be shown from the formulae for $A,B$ that both $\rightarrow 1$. It then follows from Eq.5 and Eq.8 that $\sigma_{\xi}^2=\sigma_{eff}^2/(1-\mu)$ for $\mu \alpha\leq 1$ and $\sigma_{\xi}^2=\sigma^2/(1-\mu)$ for $\mu \alpha\geq 1$. In either case, one must have $\mu\le 1$ for this limit to produce a solution to the simultaneous equations, so we conclude that $\tau\rightarrow 0$ corresponds to $\mu\le 1$, and that for these values of $\mu$ the generalization error is given by 
\begin{equation}
    GE_1(\lambda\rightarrow 0,\mu<1) = \frac{\sigma^2+\rho(1-\mu\alpha)\theta(1-\mu\alpha)}{(\sigma^2+\rho)(1-\mu)} 
\end{equation}
Notably, in this limit, $l_2$ and $l_1$ regularization yield the same results - thus, there is no difference between $l_2$ and $l_1$ penalized interpolation (see Figure 1 for numerical confirmation). This is intuitively consistent with the observation that in this limit $\hat{\rho}=1$, so that the sparsity constraint is not active - one does not expect it to have an extra beneficial effect as a result. Nevertheless, this constitutes an analytical result for the generalization error for $l_1$ penalized interpolation, and shows that the overfitting peak at $\mu=1$ also appears for the $l_1$ case. The behavior of GE to the left of the overfitting peak is identical for $l_2$ and $l_1$ penalties when $\lambda\rightarrow 0$ However, the behavior to the right of the overfitting peak is quite different for the two penalty terms, as can be seen from the considerations below and from Fig.1. 

{\bf Case 2: $\tau>0$, $\hat{\rho}\rightarrow 1/\mu$, $\implies \mu>1$} \\

This is the more interesting case, as both the nonlinear equations remain operative. To obtain analytical insight one needs to take a further limit. Within the overparametrized regime $\mu>1$, We consider $\mu\approx 1$ (slight overparametrization), and $\mu\rightarrow\infty$ (large overparametrizationn). We consider these cases separately. 

{\bf Case 2.1: $\tau$ small, $\mu \gtrsim 1$} \\
For small amounts of overparametrization, one can recover the behavior of GE by looking at the case where $\tau$ is nonzero but small compared to 1. In this case, one can proceed by expanding the LHS of Eqs 5,6 or 8,9 for small $\tau$ and retaining terms to linear order in $\tau$. Solving the resulting simultaneous linear equations for $\tau$ and substituting to obtain $\sigma_{\xi}$, one obtains 

\begin{equation}
    GE_1(\lambda\rightarrow 0,\mu\gtrsim 1) \simeq \frac{\sigma^2+\rho(1-\mu\alpha)\theta(1-\mu\alpha)}{(\sigma^2+\rho)(\mu-1)} 
\end{equation}

Thus the noise peak is recovered for $\mu\gtrsim 1$. Note that unlike the case for $\mu<1$ where we obtain an exact expression for GE, this expression is only approximately valid for $\mu\gtrsim 1$, but the approximation improves as $\mu\rightarrow 1+$ (this is numerically confirmed in Figure 1). However, the situation is quite different when $\mu>>1$ as can be seen below. 

{\bf Case 2.2: $\tau\rightarrow \infty$, $\mu \rightarrow \infty$} 

We are particularly interested in the case of large overparametrization, ie $\mu>>1$. We confine our attention to Eq.s 7,8 since for large $\mu$ and $\alpha$ fixed, eventually one would get $\mu\alpha>1$. First consider the case $\mu\rightarrow \infty $. In this case, it follows from Eq.7 that one must have $\tau\rightarrow \infty$. More precisely, setting $\hat{\rho}=\frac{1}{\mu}$ in Eq.7, in the large $\mu$ limit we get $\mu\approx e^{\frac{\tau^2}{2}}/\tau$, so that $\tau\sim\sqrt{2\log(\mu)}$ for large $\mu$. 

In this limit, one obtains from the corresponding formulae above that $A\rightarrow 0$ and $B\rightarrow 1/\sigma_{\xi}^2$. At large $\tau$,  $A(\tau)\sim e^{-\frac{\tau^2}{2}}/\tau^3$. Thus, for large $\mu$, where $\mu\approx e^{\frac{\tau^2}{2}}/\tau$, one obtains $\mu \alpha A(\tau) \sim 1/log(\mu)$, which $\rightarrow 0$ as $\mu\rightarrow\infty$ (although note that it does so slowly, only logarithmically - this will have an implication for the finite $\mu$ behavior as can be seen below). After substitution into Eq.8, one therefore obtains $\alpha\sigma_{\xi}^2=\sigma^2+\rho$, and thus
\begin{equation}
GE_1(\lambda\rightarrow 0,\mu\rightarrow \infty)=1
\end{equation}. 
Thus, in the limit $\mu\rightarrow\infty$, overparametrization ceases to be effective, and for sufficiently large overparametrization $\mu$ the inferred model can no longer generalize.

However, when $\sigma^2$ and $\rho$ are small, there is an interesting regime for values of $\mu$ where $\sigma_{\xi}^2$ is small and good generalization is possible. In this regime, $\mu,\tau>>1$ but $\tau\sigma_{\xi}<<1$. We expand the LHS of Eq.7, noting that $\hat{\rho}=1/\mu$, to obtain 
\begin{equation}
    \frac{\rho}{\alpha}(1-\sqrt{\frac{2}{\pi}}\sigma_{\xi}\tau)+\sqrt{\frac{2}{\pi}}
    \frac{\mu e^{-\frac{\tau^2}{2}}}{\tau}\approx 1
\end{equation}
This implies that $\tau\sim\sqrt{2\log(\mu)}$, provided $\rho/\alpha, \sigma_{\xi}\tau <<1$. The last condition implies that $\log(\mu)<<1/\sigma_{\xi}^2$. To satisfy Eq.8 and maintain $\sigma_{\xi}\sim\sigma$, one also requires $\rho B$ to be small. Since $B\approx 1+\tau^2$ one gets the additional condition $\tau^2\sim\log(\mu)<<1/\rho$. 

Collecting these conditions together, we find that if $1<<\log(\mu)<<1/\sigma^2,1/\rho$, and $\sigma^2,\rho/\alpha<<1$, then $GE_1\sim \sigma^2/(\sigma^2+\rho)$. In this regime, where the noise variance is small, there is significant sparsity and the degree of overparametrization is modest, the $l_1$ penalized interpolation provides much better generalization than the $l_2$ case. This can be seen numerically in Fig.1. In fact, from this figure, it would be difficult to predict that $GE_1(\mu\rightarrow\infty)=1$, however the thoretical considerations above show that this must be the case for sufficiently large $\mu$. Nevertheless, the figure shows that for large $\mu$ there can be a large difference between the generalization errors for interpolating regression with $l_1$ and $l_2$ penalties, depending on which penalty term is used. This difference is cleanly demonstrated in the limit $\sigma\rightarrow 0$ as can be seen in the next section. 

\subsubsection{Analytical expressions for GE with $l_1$ penalty when $\sigma=0$} 

This is an interesting limit since for $\sigma=0$ one can observe the algorithmic phase transition phenomena associated with $l_1$ penalized sparse regression\cite{DonohoPT}, and there is a parameter range where $GE=0$, ie there is perfect generalization (in contrast with the $l_2$ case where there is no such range). It allows for a clean demonstration of one of the main points of the paper, that the region of good generalization is demarcated by $\mu\alpha=1$ rather than by $\mu=1$ (the interpolation point). The small $\sigma$ case considered earlier can be qualitatively understood in the $\sigma\rightarrow 0$ limit, and an important phenomenon is observed, namely that there is a regime of large overparametrization where $GE_1=0$. We separately consider the regimes $\mu<1$ and $\mu>1$.

{\bf Case 1}: $\mu<1$ (underparametrized regime): 

In this case, $GE_1$ can be obtained by setting $\sigma=0$ in Eq.9, 

\begin{equation}
  GE_1(\sigma=0,\lambda\rightarrow 0,\mu<1) = \frac{(1-\mu\alpha)\theta(1-\mu\alpha)}{1-\mu}   
  \nonumber
\end{equation}

If $\alpha<1$, then the theta function is not operative in this regime, and one obtains the noise peak as $\mu$ approaches 1 from the left. Although there is no additive noise, the underparametrization produces an effective noise with strength $\sigma_{eff}^2=\rho(1-\mu\alpha)$, and this leads to the noise peak. However, if $\alpha>1$, then $GE_1$ goes to zero at $\mu=\frac{1}{\alpha}<1$, and stays zero in the regime $\frac{1}{\alpha}<\mu<1$. Also, there is no noise peak for $\alpha>1$. Note, as before, that in this underparametrized regime, $GE_2(\lambda\rightarrow 0)=GE_1(\lambda\rightarrow 0)$.

{\bf Case 2}: $\mu>1$ (overparametrized regime): In this regime, $l_2$ and $l_1$ penalized interpolation can produce dramatically different generalization behavior. 

We first consider the $l_2$ case for $\sigma=0$, $\mu>1$. In this case, it follows from the formula in section 3.3 that 

\begin{equation}
    GE_2(\sigma=0,\lambda\rightarrow 0,\mu>1) = 1-\frac{1}{\mu}+(1-\mu\alpha)\theta(1-\mu\alpha)\bigl[\frac{1}{\mu-1}+\frac{1}{\mu}\bigr]
    \nonumber
\end{equation}

For $\alpha<1$, the noise peak is present, and the generalization error reaches a minimum value of $GE_2(\sigma=0,\lambda\rightarrow 0,\mu=\frac{1}{\alpha}) = 1-\alpha$, after which it increases as $1-\frac{1}{\mu}$ with increasing $\mu$. For $\alpha>1$, $GE_2(\sigma=0,\lambda\rightarrow 0,\mu\alpha>1) = 1-\frac{1}{\mu}$. In either case, $GE_2(\sigma=0,\lambda\rightarrow 0,\mu>1)>0$. 

On the other hand, we will show below that if $\alpha>\alpha_c(\rho)$ (to be defined below), then 
\begin{eqnarray}
GE_1(\sigma=0,\lambda\rightarrow 0,\mu)&=&0\nonumber\\
\text{for}\quad\frac{1}{\alpha}<&\mu&<\mu_c(\frac{\alpha}{\rho})\approx \sqrt{\frac{\pi\alpha}{2\rho}}\operatorname{exp}\bigl[{\frac{\alpha}{2\rho}}\bigr] \nonumber
\end{eqnarray} 

$GE_1>0$ for $\mu>\mu_c(\frac{\alpha}{\rho})$ and $GE_1(\sigma=0,\lambda\rightarrow 0,\mu\rightarrow\infty)=1$. When $\mu_c$ is large (which is the case when $\frac{\rho}{\alpha}<<1$), there is a sizeable region of large overparametrization where $GE_1(\lambda\rightarrow 0)=0$ but $GE_2(\lambda\rightarrow 0)\approx 1$ (consistent with Fig.1). This "gap" region differentiates $l_2$ and $l_1$ penalized interpolating regression and shows that good generalization is not a property of regularized interpolation per se, but depends strongly on the method of interpolation. 

In order to show the existence of the gap region, and derive the formula for $\mu_c$, we need to study the Eq.s 4-8 after setting $\sigma=0$.  

{\bf Case 2.1}: $\mu>1$ and $\alpha<1$. 

In this case, there is a noise peak even for $\sigma=0$ due to the effective noise. For $\mu\gtrsim 1$, from Eq.10 the noise peak shape matches that of $l_2$ penalized interpolation, $GE_1\sim \frac{1-\mu\alpha}{\mu-1}$. However, as $\mu$ increases and $\sigma_{eff}^2=\rho (1-\mu\alpha)\rightarrow 0$, the effective noise vanishes. When $\sigma_{eff}=0$ Eq.s 4,5 are identical to the corresponding equations for ordinary $l_1$ penalized regression \cite{Ramezanali2019} and we expect a continuous phase transition where $\sigma_{\xi}\rightarrow 0$ as $\mu\rightarrow \frac{1}{\alpha}$. 

From Eq.6, since $\mu>1$, we cannot have $\tau=0$, and as we approach the transition from the left, $\sigma_{\xi}>0$. Thus to satisfy Eq.6 as $\lambda\rightarrow 0$ we have $\hat{\rho}=\frac{1}{\mu}$. Expanding Eq.s 4,5 to leading order in $\sigma_{\xi}$, which is assumed small, and setting $\mu=\frac{1}{\alpha}-\delta\mu$, we obtain the equations (Note that $A(\tau)$ is defined below Eq.6)
\begin{eqnarray}
    \alpha+\alpha^2\delta\mu&=&F_1(\tau,\rho)-\sqrt{\frac{2}{\pi}}\rho\tau\sigma_{\xi}+O(\sigma_{\xi}^2,\delta\mu^2)\\
    \alpha+\alpha^2\delta\mu&=&\frac{\rho\alpha\delta\mu}{\sigma_{\xi}^2}+F_2(\tau,\rho)-\frac{2}{3}\sqrt{\frac{2}{\pi}}\sigma_{\xi}\tau(3+\tau^2)+O(\sigma_{\xi}^2,\delta\mu^2) \\
    F_1(\tau,\rho)&=&1-(1-\rho) \operatorname{Erf}(\frac{\tau}{\sqrt{2}}) \nonumber\\
    F_2(\tau,\rho)&=&(1-\rho) A(\tau)+\rho(1+\tau^2)\nonumber
\end{eqnarray}
From Eq.14 it can be seen that as $\sigma_{\xi}\rightarrow 0$, one must have $\frac{\delta\mu}{\sigma_{\xi}^2}\rightarrow c_1$ where $c_1$ is a finite non-negative constant (possibly zero). Thus we obtain the following equations that must be satisfied for the $\sigma_{\xi}\rightarrow 0$ limit to exist: 
\begin{eqnarray}
    \alpha&=&F_1(\tau,\rho)\\
    \alpha&=&\rho\alpha c_1+F_2(\tau,\rho)
\end{eqnarray}
$F_2(\tau,\rho)$ is an analytic function of $\tau$ and it can be verified that it is positive, with a minimum at a critical value of $\tau=\tau_c(\rho)$ given by the equation $F_1(\tau,\rho)=F_2(\tau,\rho)$. The value of $\alpha$ at this critical point at which $F_2$ is a minimum is given by $\alpha_c(\rho)=F_1(\tau_c(\rho),\rho)$. Thus, for the equations to be satisfied, we must have $\alpha\geq \rho\alpha c_1 + \alpha_c(\rho)$, where $c_1\geq 0$. It follows that no solution with $\sigma_{\xi}\rightarrow 0$ exists if $\alpha< \alpha_c(\rho)$. If $\alpha>\alpha_c(\rho)$, it also follows that $c_1>0$, thus demonstrating that $\sigma_{\xi}^2\propto \delta\mu$ and also $GE_1\propto\delta\mu$ as $\mu\rightarrow \frac{1}{\alpha}-$. The proportionality constant can be worked out to be 
\begin{eqnarray}
    GE_1(\sigma=0,\lambda\rightarrow 0,\mu\rightarrow \frac{1}{\alpha}-)&=&\frac{\alpha}{\alpha-F_2(\tau_0,\rho)}\delta\mu+O(\delta\mu^2)\nonumber\\
    \tau_0&=&\operatorname{Erf}^{-1}\bigl(\sqrt{2}\frac{1-\alpha}{1-\rho}\bigr)\nonumber
\end{eqnarray}
Here we use the notation $\mu\rightarrow \frac{1}{\alpha}-$ to denote that $\mu$ approaches $\frac{1}{\alpha}$ from below. Note that as $\tau_0\rightarrow \tau_c(\rho)$ the slope diverges. This is due to the fact that exactly at $\alpha=\alpha_c(\rho)$, $\sigma_{\xi}^2$ goes to zero as a power of $\delta\mu$ that is smaller than 1. At $\alpha=\alpha_c(\rho)$ from Eq.14 one has $\frac{\delta\mu}{\sigma_{\xi}^2}\sim \sigma_{\xi}$, so that $\sigma_{\xi}\sim(\delta\mu)^{\frac{1}{3}}$ and $GE_1(\mu\rightarrow \frac{1}{\alpha}-)\propto (\delta\mu)^{\frac{2}{3}}$

This completes analysis of the case $\mu<\frac{1}{\alpha}$. For $\mu>\frac{1}{\alpha}$, one no longer has the effective noise term. There are two cases: $\alpha<\alpha_c(\rho)$ and $\alpha > \alpha_c(\rho)$. In the former case, $\sigma_{\xi}$ remains finite at $\mu=\frac{1}{\alpha}$ and from an examination of the continuity of Eq.4,5 with Eq.7,8 at $\mu=\frac{1}{\alpha}$, continues to remains finite for $\mu>\frac{1}{\alpha}$. In the latter case however, one has a non-trivial solution with $\sigma_{\xi}=0$, given by the equations 
\begin{eqnarray}
    \hat{\rho}&=&F_1(\tau,\frac{\rho}{\mu\alpha})\\
    \frac{1}{\mu}&=&F_2(\tau,\frac{\rho}{\mu\alpha})
\end{eqnarray}
Note that since $\sigma_{\xi}=0$, the relation $\hat{\rho}\mu=1$ no longer has to hold, but one must still have $\mu\hat{\rho}\leq 1$. Solving these equations simultaneously for $\hat{\rho},\tau$, one obtains $\hat{\rho}(\mu,\frac{\rho}{\alpha})$ and $\tau(\mu,\frac{\rho}{\alpha})$ as functions of $\mu,\frac{\rho}{\alpha}$. 

It is important to note that just after the transition to $\sigma_{\xi}=0$ at $\mu=\frac{1}{\alpha}$, the RHS of Eq.18 is different from the RHS of Eq.16 since the term $\rho\alpha c_1$ is missing. This has a significant consequence. Note that $F_2(\tau,\rho)$ has a minimum value of $\alpha_c(\rho)$ at $\tau=\tau_c(\rho)$. Since $\alpha>\alpha_c(\rho)$, one notes that there are two solutions $\tau_1<\tau_c(\rho)<\tau_2$ to Eq.18. Further, examination of Eq.s 16 and 18 shows that at $\mu\alpha=1$, the solution $\tau_0$ to Eqs. 15,16 satisfies $\tau_0>\tau_1$. Since $F_1(\tau,\rho)$ is a decreasing function of $\tau$, it follows that $F_1(\tau_1,\rho)>F_1(\tau_0,\rho)=\alpha$. Thus, taking the solution $\tau_1$ to Eq.s 17,18 would result in $\hat{\rho}>\alpha=\frac{1}{\mu}$ just to the right of the transition point at $\mu\alpha=1$. This is not permissible. Therefore just to the right of the transition point at $\mu\alpha=1$ there is a jump, and one must adopt the value of $\tau_1>\tau_c(\rho)$. This will result in a value of $\hat{\rho}<\frac{1}{\mu}$. Thus the number of non-zero estimated parameters (for infinitesimally small $\lambda$) jumps from $\frac{1}{\mu}=\alpha$ to a smaller value as $\mu$ crosses the transitional  value $\frac{1}{\alpha}$.  

As $\mu$ increases however, $\frac{1}{\mu}$ decreases and eventually one reaches the point at which $\hat{\rho}$ again reaches the value $\frac{1}{\mu}$. The point $\mu_c$ at which this happens is obtained by setting $\hat{\rho}=\frac{1}{\mu_c}$ in Eq.17. This gives rise to the equation 

\begin{equation}
    \frac{1}{\mu_c}=\alpha_c(\frac{\rho}{\mu_c \alpha})
\end{equation}

Note that this equation has the form $\alpha_{eff}=\alpha_c(\rho_{eff})$ with $\alpha_{eff}=\frac{1}{\mu_c}$ and $\rho_{eff}=\frac{\rho}{\mu_c \alpha}$, with $\alpha_{eff}/\rho_{eff}=\frac{\alpha}{\rho}$. This makes sense, since $\alpha_{eff}=m/p$ (instead of $m/n$) and $\rho_{eff}=\rho n/p$. Thus $p$ (the number of fitting parameters) plays the role of $n$ (the number of parameters in the generative model which could be potentially nonzero). This equation helps explain why there is a transition when $\mu$ is large: as $\mu\rightarrow\infty$, both $\alpha_{eff}$ and $\rho_{eff}$ become small while remaining proportional to $\alpha/\rho$ which is fixed. As $\mu$ increases, ($\alpha_{eff},\rho_{eff}$) moves towards the origin along the straight line $\alpha_{eff}=\frac{\alpha}{\rho}\rho_{eff}$. This line intersects the curve $\alpha_{eff}=\alpha_c(\rho_{eff})$, giving the transition point when one goes from the "perfect recovery" regime to the regime where recovery is not possible in the LASSO problem.

\begin{figure}
  \centering
\includegraphics[width=0.5\linewidth]{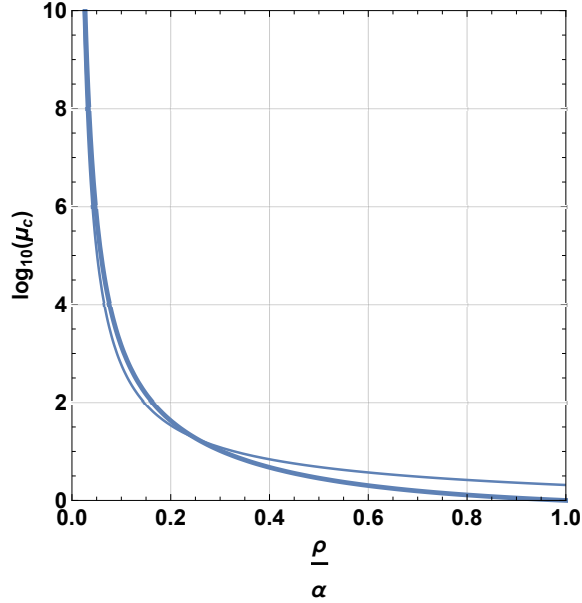}
    \caption{Critical value of overparametrization $\mu_c$ as a function of $\frac{\rho}{\alpha}$ (thick line) together with the approximate form $\mu_c\approx\sqrt{\frac{\pi\alpha}{2\rho}}e^{\frac{\alpha}{2\rho}}$ (thin line). Note that the plot is semi-logarithmic in base 10. 
}
\end{figure}

The equations giving the transitional value $\mu_c$ can be obtained by some algebra from Eqs 17,18 after setting $\hat{\rho}=\frac{1}{\mu_c}$, and are given by 
\begin{eqnarray}
    \frac{\rho}{\alpha}&=& 1-\sqrt{\frac{\pi}{2}}\tau e^{\frac{\tau^2}{2}}\operatorname{Erfc}(\frac{\tau}{\sqrt{2}})
    \nonumber\\
    (\mu_c-1)\frac{\rho}{\alpha}&=& \frac{\pi}{2} \tau \operatorname{exp}\bigl[{\frac{\tau^2}{2}}\bigr]\nonumber
\end{eqnarray}
This explicitly demonstrates the logarithmic relationship between $\tau$ and $\mu_c$ (for large $\mu_c$) that was utilized earlier in section 3.3.3. These equations give $\mu_c$ as a function of $\frac{\rho}{\alpha}$ in parametric form (see Figure.4, thick line). A good approximation to $\mu_c$ can be obtained by taking the small $\rho$ limit, and is given by (see Fig.4, thin line) 
\begin{equation}
    \mu_c\approx\sqrt{\frac{\pi\alpha}{2\rho}}\operatorname{exp}\bigl[{\frac{\alpha}{2\rho}}\bigr]
\end{equation}
$\mu_c$ can be large even for modest values of $\rho$. For the parameters chosen in Fig.1 ($\rho=0.2,\alpha=0.8$), one gets $\mu_c=18.7$. Thus, for these parameter choices, one can overparametrize upto 18 times, without entering into the regime where $GE_1>0$ in the interpolating regime for $\sigma=0$ (a behavior that qualitatively carries over to small $\sigma$). On the other hand, in this regime, the generalization error of interpolating regression with $l_2$ penalty $GE_2$ gets close to its asymptotic value of 1. 

Finally, for $\mu>\mu_c$, $\sigma_{\xi}$ rises from zero linearly in $\delta\mu=\mu-\mu_c$. This can be seen by expanding Eq.7,8 in $\sigma_{\xi}$ (similar to Eq.13,14). In this case the effective noise term $\rho(1-\mu\alpha)$ is absent, and one simply obtains $\sigma_{\xi}\propto (\delta\mu)^2$, leading to $GE_1\propto (\delta\mu)^2 $. Arguments presented in section 3.3.3 show that $GE_1\rightarrow 1$ as $\mu\rightarrow \infty$. 

\subsection{$l_1$ Risk Curves: Derivation}

The derivation of the generalization error for $l_1$ regularized regression in the high-dimensional asymptotic limit is somewhat involved and can only be sketched here. Many approaches can be brought to bear on this problem, including the replica and cavity approaches from statistical physics\cite{Kabasjima99,Kabashima09,Ganguli10,KrzakalaCS} and the message-passing/belief propagation approaches\cite{DonohoOMP,DonohoAMP,Montanari-book}. These approaches produce the same results. Here we use the cavity approach. 

Let $u_i=\hat{\beta_i}-\beta_i$ be the error made in estimating the $i^{th}$ parameter using penalized mean square optimization. In the cavity approach, the generative parameter $\beta_i$ corresponding to a given site $i$ is held fixed, while the other generative parameters as well as the design matrix are allowed to vary according to the corresponding probability distributions. This gives rise to a stochastic equation satisfied by $u_i$, where the stochasticity is the effect of the quantities in the generative model that are varying from realization to realization. 

In the large $n,m,p$ limit with the ratios fixed as before, the stochastic behavior simplifies and can be characterized by a single stochastic variable $\xi_i$ for each site, with $\xi_i$ being i.i.d normally distributed. The variance of $\xi_i$ has to be determined self-consistently. These equations are as follows ($\xi$ is normally distributed): 

{\it Underspecified case $\mu \alpha < 1, p < n$}: all $u_0$ residuals are equivalent, and satisfy a stochastic equation of the form below \cite{ramezanali2015cavity,Ramezanali2019}:

\begin{eqnarray}
        u_0 + \lambda^{'} V^{'}(\beta_0 + u_0) &=& \xi   \nonumber\\
        V(\xi) &=& \frac{1}{\alpha}(\sigma_{eff}^2 + \mu\alpha V(u_0))\nonumber\\
        \lambda^{'} &=& \lambda/(\alpha c) \nonumber\\ 
        c &=& \frac{1}{m} Tr E_X(1 - X\chi X^+) \nonumber
\end{eqnarray}
The matrix $\chi$ is given by 
\begin{equation}
        [\chi^{-1}]_{ij} = [X^+X + \lambda~diag [V^{''}(\beta_i + u_i^0)]]_{ij}\nonumber\\
\end{equation}

Here $X$ is the $m\times p$ design matrix and $\chi^{-1}$ has on its diagonals the second derivatives of the regularization penalty term. For this equation to be applied to the $l_1$ case, $V(\beta)=|\beta|$ has to be smoothed slightly around $\beta=0$ to make it second differentiable, and a limiting procedure applied where the smoothing is removed after computation of the expectation over $X$ (this is equivalent to a sub-derivative procedure). The expectation over $X$ can be computed using diagrammatic perturbation theory for random matrices in the large $n$ limit. This yields the crucial equation Eq.6 defining the effective parameter $\lambda^{'}$:\\
\begin{eqnarray}
    c&=&(1-\mu\hat{\rho}) \nonumber\\ 
    \frac{\lambda}{\alpha} &=& (1-\mu\hat{\rho})\lambda^{'}   \nonumber 
\end{eqnarray}

For convenient parametrization, we set $\lambda^{'}=\tau\sigma_{\xi}$, as this simplifies the notation, and obtain Eq.6. The variable $\tau$ appears in the equations for computing the variance $\sigma_{\xi}^2$ self-consistently. The generalization error is given by \\
\begin{equation}
GE_1=\frac{\sigma_{eff}^2 + \mu\alpha V(u_0)}{\sigma^2+\rho}=\frac{\alpha \sigma_{\xi}^2}{\sigma^2+\rho}
\end{equation}

Details of the derivation of the stochastic equations, and the random matrix computations, will not be presented here, but follow the cavity approach applied to $l_1$ penalized regression in \cite{ramezanali2015cavity,Ramezanali2019}. 

Equations 4,5 are obtained from the stochastic equation for $u_0$ by considering the ranges of $\xi$ for which $u_0$ is zero (thus giving an equation for $\hat{\rho}$), and computing the variance of $u_0$ together with the self-consistency condition above for $\sigma_{\xi}^2$. These equations contain integrals over the distribution $\pi(\beta)$, which may be analytically computed for the case that $\pi(\beta)$ is a normal distribution. 

The corresponding algebra is straightforward but tedious will not be detailed here. We will only note one important step in the derivation relating to the range of $\xi$ for which $u_0=0$. Consider the case $\beta_0=0$, and as suggested previously, smooth the corner of $|\beta|$ so that $V^{'}(u_0)$ is a slightly smoothed step  function. Then for $-\lambda^{'}<\xi<\lambda^{'}$ $u_0\approx 0$. Thus, by separating ranges for $\xi$ for which $u_0$ remains pinned to zero from ranges where it differs from zero, the equations for $\hat{\rho}$ and $V(u_0)$ can be written down.

{\it Overspecified case $\mu \alpha > 1, p > n$}: two classes of residuals $u_1$ and $u_2$ have to be considered, corresponding to the parameters which belong to the generative model and can have nonzero values, and the extra parameters added during fitting which never have non-zero values in the generative model. Below, $\xi_1$ and $\xi_2$ have uncorrelated normal distributions with the same variance,

        \begin{eqnarray}
        u_1 + \lambda^{'} V^{'}(\beta_1 + u_1) &=& \xi_1 \nonumber\\
        u_2 + \lambda^{'} V^{'}(\beta_2 + u_2) &=& \xi_2 \nonumber\\
        \beta_1 &\sim& (1-\rho)\delta(\beta_1) + \rho\pi(\beta_1) \nonumber\\ \beta_2 &=& 0\nonumber\\
        V(\xi_1) &=& V(\xi_2) = V(\xi)\nonumber\\
    V(\xi) &=& \frac{1}{\alpha} [\sigma^2 + V(u_1) + (\mu \alpha - 1)V(u_2)] \nonumber
\end{eqnarray}

Equations 7,8 follow from the above equations following the same procedure as for the underspecified case. 

\section*{Acknowledgement} This work was supported by the Crick-Clay Professorship (CSHL) and the H N Mahabala Chair Professorship (IIT Madras). Help from Dr Jaikishan Jaikumar in preparing the figures shown in the paper is gratefully acknowledged. 

\bibliography{main}
\bibliographystyle{plain}

\end{document}